# Applying Gene Expression Programming for Solving One-Dimensional Bin-Packing Problems

Najla Akram Al-Saati
dr.najla_alsaati@uomosul.edu.iq
*College of Computer Sciences and Mathematics/University of Mosul*

## Abstract

This work aims to study and explore the use of Gene Expression Programming (GEP) in solving on-line Bin-Packing problem. The main idea is to show how GEP can automatically find acceptable heuristic rules to solve the problem efficiently and economically. One dimensional Bin-Packing problem is considered in the course of this work with the constraint of minimizing the number of bins filled with the given pieces. Experimental Data includes instances of benchmark test data taken from Falkenauer (1996) for One-dimensional Bin-Packing Problems. Results show that GEP can be used as a very powerful and flexible tool for finding interesting compact rules suited for the problem. The impact of functions is also investigated to show how they can affect and influence the success of rates when they appear in rules. High success rates are gained with smaller population size and fewer generations compared to a previous work performed using Genetic Programming.

<div dir="rtl">

**تطبيق البرمجة بالتعبير الجيني في حل مسائل تعليب الصناديق**

نجلاء اكرم الساعاتي

كلية علوم الحاسوب والرياضيات / جامعة الموصل

تاريخ استلام البحث: 2012/9/24     تاريخ قبول البحث: 2013/1/30

**الخلاصة**

يهدف هذا البحث الى دراسة واستكشاف استخدام طريقة البرمجة بالتمثيل الجيني في حل مسالة التعليب (ملأ العلب او تعليب الصناديق) بالطريقة الفورية المباشرة (on-line). وترتكز الفكرة الاساسية على توضيح الكيفية التي تتمكن فيها طريقة (GEP) من ايجاد قواعد حدسية مقبولة وجيدة بشكل تلقائي لأجل حل المسالة بشكل كفوء واقتصادي. تم اخذ مسالة التعليب ذات البعد الواحد في مسار هذا البحث بوجود قيد تقليل عدد الصناديق او العلب التي تملأ بالقطع المعطاة. تتضمن البيانات التجريبية حالات نموذجية قياسية مأخوذة من العالم Falkenauer في عام 1996 لمسائل التعليب احادية البعد. تبين النتائج انه بالإمكان استخدام طريقة (GEP) كأداة قوية جدا ومرنة لإيجاد قواعد محكمة ومفيدة تلائم المسالة. تناول البحث ايضا استقصاء تأثير الدوال المستخدمة في القواعد لإظهار كيفية تأثيرها ونفوذها على نسب النجاح حين تشارك في تكوين القاعدة او القانون. وقد تم استحصال نسب نجاح عالية جدا باستخدام كثافة سكانية اقل وعدد اصغر من الاجيال عند مقارنتها مع اعمال سابقة استخدمت فيها طريقة البرمجة الجينية (Genetic Programming).

***الكلمات المفتاحية****: البرمجة بالتمثيل الجيني ، مسالة التعليب ، البرمجة الجينية.*

</div>

## 1. Introduction

Artificial intelligence has gone a long way in solving various optimization problems; its tools have had an immense role in finding good and acceptable solutions for such problems. New AI algorithms that are inspired by nature keep emerging out periodically; all competing to demonstrate improved performance and enhanced applicability.

Bin-Packing is a well-known classical optimization problem that belongs to the more general class of Packing Problems. Methods that seek solutions for such problems usually intend to find the best packing strategy by determining the best way to pack the items so that they all fit into a minimum number of storage bins. The application of this sort of problem to real world situations can be of a great practical significance, and it can be found in many appropriate areas such as transportation, production, industrial regions and many more. It is found to be very useful in many circumstances, like packing up containers, loading trucks with weight capacity, even constructing file backup in removable media [3]. Bin-Packing is also of an essential theoretical importance, being used as an early proving base for many of the classical approaches to analyze the performance of approximation algorithms [7].

The Bin-Packing problem has been studied in computer science ever since 1970 [31]. It comes in many variations such as linear packing, packing by size, weight, or cost, and many more, it can also be of multi-dimensions. Bin-Packing has a special importance in view of the fact that it is NP-hard [7]; a lot of investigations have been made throughout the literature to fine good enough heuristics that can solve the problem. New algorithms and hyper heuristics are used with the goal of minimizing the number of bins used in the process.

This work introduces a technique inspired by the idea of building computer programs capable of developing new interesting heuristics. This is done through the use of Gene Expression Programming (GEP) as a hyper heuristic to find the best packing strategy or heuristic to fill the bin containers, the idea is to discover a solution process rather than to solve an instance of problem. GEP was used to solve many difficult optimization problems since 2002 when it was first introduced by Ferreira. It outperforms GP due to the fact that its individuals are encoded in linear strings of fixed length (genome) and are then expressed as non-linear entities of various sizes and shapes (phenome or expression trees). Chromosomes can be of one or more genes that are linked by a function to evaluate the fitness of the chromosome as a whole individual.

In this work, *on-line bin-packing* problem is considered, i.e., the number of pieces is not known in advance nor are their sizes. The proposed *one-dimensional* bin-packing method packs the pieces into the bins in their arrival order, where pieces are not be unpacked once they have been placed in a bin. This situation is likely to arise in the real world. As for GEP, single-gene and multi-gene chromosomes are considered in this work.

## 2. Previous Work:

There are many approaches in the literature involving the solution of the bin packing problem, some introduce new techniques, and others make new variations and even add new constraints to the problem. Many heuristics and hyper-heuristics were developed over the last years, evolutionary algorithm such as genetic algorithms and genetic programming were used to produce good quality solutions to the problem.

Falkenauer (1996) [9] viewed the packing of items in bins as the process of grouping, and he introduced the 'Grouping Genetic Algorithm' (GGA) which is a Genetic Algorithm that is heavily modified to suit the structure of grouping problems.

Fleszar and Hindi, (2002) [14] suggested a few new heuristics to solve the Bin-Packing problem, the most effective of these heuristics are those based on the *variable neighborhood search* (VNS) meta-heuristic [18].

Alvim et al. (2004) [1] proposed a hybrid improvement procedure for the bin packing problem. They showed that their heuristic has several features such as the use of lower bounding strategies, the generation of initial solutions by reference to the dual min-max problem, the use of load redistribution based on dominance, differencing, and unbalancing; and the inclusion of an improvement process utilizing tabu search.

Jing et al. (2006) [20] worked out a hybrid genetic algorithm for bin packing problems based on item sequencing. They used a simple GA to search the solution of bins sequence, and the next fit algorithm to pack the sequenced items into the bins sequence obtained.

Liu et al. (2008) [25] solved the multi-objective bin-packing problems with evolutionary particle swarm optimization for two dimensional problems.

Ant algorithms were used in 2009 to solve the problem by Benmohamed and Yassine [4]; they presented an approach of resolution combining optimization by colony of ants (ACO) and the heuristic method IMA to resolve this NP-hard problem. In addition they treated the case of objects with a free orientation of 90◦ for two-dimensional bin-packing.

Muritiba, et al. (2010), treated the problem with an exact approach, based on a set covering formulation solved through a branch-and-price algorithm [29]. While Khanafer *et al.* proposed a general framework for deriving new data-dependent dual feasible functions in order to describe new lower bounds for the problem [21].

Recently in 2011, Maiza and Radjef introduced a couple of heuristics to solve the one-dimensional problem with conflicts [27]. The conflicts are represented by a graph whose nodes are the items, and adjacent items cannot be packed into the same bin. They proposed an adaptation of minimum bin slack heuristic with a combination of heuristics based on using the classical bin-packing methods to pack items of maximal-stable-subsets (MSS).

Layeb and Boussalia (2012) [22], developed a novel GRASP (Greedy Randomized Adaptive Search Procedure) algorithm for solving the bin packing problem. Shortly after that, and in the same year, also in (2012) Layeb and Chenche presented a novel quantum inspired cuckoo search algorithm to solve the problem [23].

As for approximate solutions to the problem, the fastest heuristics are the well-known First-Fit Decreasing (FFD) and Best-Fit Decreasing (BFD) greedy algorithms [28]. A number of heuristics usually used to solve the Online Bin Packing Problem are as follows:

- *Best Fit*: Put the piece in the fullest bin that has room for it, opens a new bin if it doesn't fit in any existing bin.
- *Worst Fit*: Put the piece in the emptiest bin that has room for it, open a new bin if it doesn't fit in any existing bin.
- *Almost Worst Fit*: Put the piece in the second emptiest bin if that bin has room for it, open a new bin if it doesn't fit in any open bin.
- *Next Fit*: Put the piece in the right-most bin, open a new bin if there isn't room for it.

*Najla A. Al-Saati*

- *First Fit*: Put the piece in the left-most bin that has room for it, open a new bin if it doesn't fit in any open bin [25].

Gene Expression Programming (GEP), on the other hand, was applied to many real world problems since it was introduced, for example, the symbolic regression, sequence induction, block stacking problems and the density-classification problem. Boolean concept learning was also investigated by Ferreira [10] such as the 11-multiplexer and the GP rule problem. In addition, she designed Neural Networks using GEP. [13] Furthermore Zhou et al. used GEP to evolve classification rules [32].

## 3. Bin-Packing Problem

Bin-Packing problem is about packing items of different sizes into the smallest possible number of bins all of a unit size. This type of problem is a combinatorial problem known to be NP-hard [7], so there is no known optimal algorithm running in polynomial time to solve it. For that reason, artificially intelligent techniques were used to yield good enough, near optimal solution to the problem.

There are two types of the classical bin-packing problems: on-line and off-line. In On-line problems, the items arrive one by one, so the algorithm can process its input piece by piece serially. Off-line problems, on the contrary, have all the items received by the algorithm from the beginning. No matter on-line or off-line, solving procedures always access the bin in an arbitrary order. [16] In this work, online problems are considered.

There is also another type of this problem; it is the *on-line bin-packing with rejection* in which the algorithm has the chance of rejecting some items. Here, the loss function is the sum of the used bins number and the rejected items' costs [19].

Some variants of bin-backing problems (both on-line and off-line) abandon the assumption that the algorithm has access to the bins in arbitrary order. *Sequential Bin-Packing* introduces a more restricted version where items arrive one by one (just like in the on-line problem) but in each round the algorithm has only two possible choices: assign the given item to the (only) open bin or to the "next" empty bin (the new open bin), and items cannot be assigned anymore to closed bins [16].

This work investigates one-dimensional on-line Bin-Packing problem, in which there is an endless supply of bins, each of capacity *C*, and a set of *n* integer-size pieces or items that must be packed in these bins without exceeding their capacity [30]. This is shown in Eq. (1)

$$\sum_{i \in bin_k} s_i \leq C \quad \ldots\ldots\ldots\ldots\ldots\ldots\ldots\ldots\ldots\ldots\ldots\ldots\ldots\ldots\ldots..(1)$$

Where
- $s_i$ is the size of item i and $s_i > 0$ for all i.
- i is the item number from 0 to n.
- k is the bin number
- C is the capacity of bins (a positive number)

The problem must be solved with the constraint that the set of items must be packed into the smallest possible number of bins. If M is the minimal number of bins needed, then,

$$M \geq \left\lceil \left( \sum_{i=1}^{n} s_i \right) / C \right\rceil \quad \ldots\ldots\ldots\ldots\ldots\ldots\ldots\ldots\ldots..(2)$$

In algorithms that start new bins only when needed, the number of used bins is ≥ M and < 2M. Since using 2M bins or more would mean that there are two bins whose combined contents is less or equal C, and they could be combined into one bin [30].

## 4. Hyper-Heuristics Vs. Meta-Heuristics:

Heuristic rules have been very widely used to sort out practical problems in operational research, and due to their NP-hard nature, exhaustive search is often computationally easier said than done. Over the last two decades, Meta-heuristics have had a major role at the crossing point of Artificial Intelligence and Operational Research; they were used for a wide and diverse range of application areas and have powerfully influenced the advance of modern search technologies. The applications of meta-heuristics can be found in many assorted areas such as scheduling, data mining, stock cutting, medical imaging and bio-informatics, and many others [15].

The idea of hyper-heuristics was first originated by Denzinger et al.(1997) [8], but was used to describe a procedure that selects and combines a number of Artificial Intelligent methods. The exact term, was first encountered by Burke, Kendall and Soubeiga (2003) [6]. The term hyper-heuristic has been defined to generally depict the process of using meta-heuristics that can choose meta-heuristics to solve a problem. In other words, a hyper-heuristic chooses a sequence of heuristics that best improves the solution. Hyper heuristics are cheaper and easier to implement than problem specific special purpose methods [15].

The hyper-heuristic Evolutionary approach deals with a population of heuristics over a number of generations to evolve these heuristics. They are usually problem independent and are easy to use. As a matter of fact, the need for hyper-heuristic arise when presented with a new instance of a problem, it is frequently not instantly recognizable to decide the appropriate method to apply.

GEP acts as a hyper-heuristic; it selects among a number of low level building blocks (function and terminal sets) to form a heuristic that acts satisfactorily in solving a problem.

## 5. Gene Expression Programming (GEP)

GEP is an approach that utilizes a genotype/phenotype system with multi-gene chromosomes which are encoded as expression trees (ETs) and linked by certain linking relations [11]. The chromosomes that are named *K-expressions* are composed of genes structurally arranged as a head and a tail. These fixed-length genes flexibly handle ETs of different shapes and sizes. The genotype/phenotype mapping of GEP guarantees the feasibility of phenotypes and results in unconstrained search of genotypes. The bottom line is that GEP combines the advantages of both GAs and GP; therefore, GEP can be used for conventional applications of GAs and GP [32].

The flowchart of a gene expression algorithm (GEA) is shown in Figure (1). The procedure starts by randomly generating an initial population of chromosomes. After that the chromosomes are expressed and assigned a fitness value. These individuals undergo selection depending on their fitness to reproduce with modification new offspring's which, in their turn, follow the same procedure. The process is repeated either for a predefined number of generations or until an acceptable solution is found.

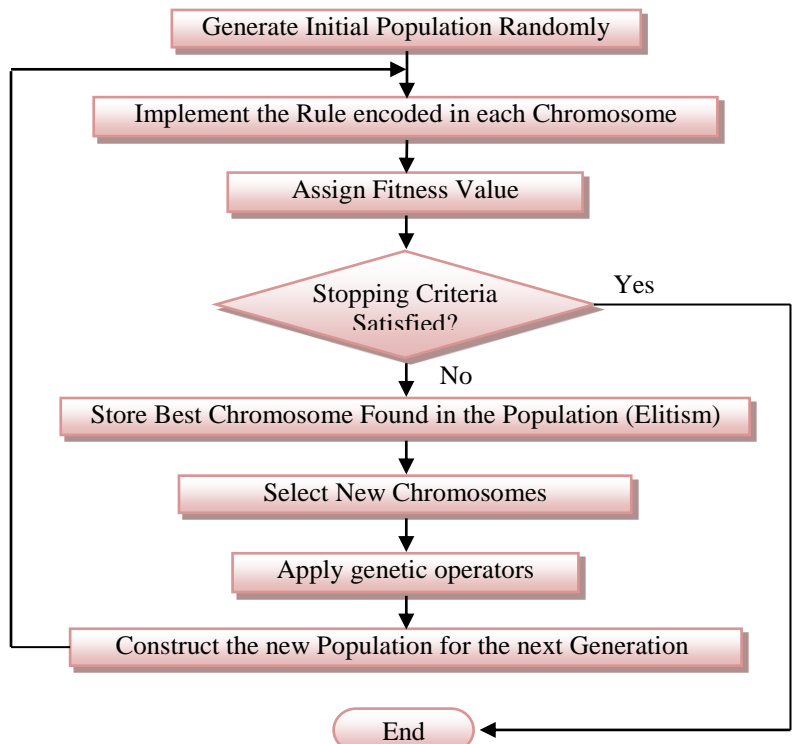

**Figure (1) Flowchart of a Gene Expression Algorithm**

GEP's general components are the function set, terminal set, fitness function, control parameters, and stopping criteria. Fixed-length strings are used to represent the individuals; these are later expressed as parse (expression) tree of different sizes and shapes to be submitted to the fitness function for evaluation [11].

The genes in GEP contain two parts: a head and a tail. The head holds symbols representing the functions and terminals, while the tail includes terminal symbols only. Usually the head's length (h) is defined, but the length of the tail (t) is computed as in the Eq. (3) below [10]. The tail is a function of the head (h) and (n) which is the maximum number of arguments taken by a function in the function set.

$t = h * (n - 1) + 1$ ……………………………………………………………….(3)

The general representation of chromosomes is shown in Figure (2); it shows a single-gene chromosome. GEP commonly uses multi-gene chromosomes that are connected by a linking function as in Figure (3).

The structure of genes is best viewed as an open reading frame (ORF). An ORF (coding sequence) of a gene begins biologically with the "**start**" codon, continues with the amino acid codons, and ends at a termination codon. In GEP the "**start**" point is always the first position of a gene, but the **termination** point is not always the last position [12]. Figure (2) depicts this idea by example.

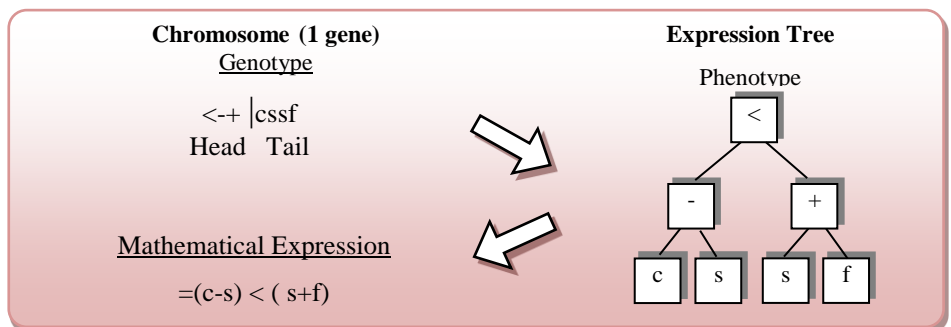

**Figure (2) Representation in GEP**

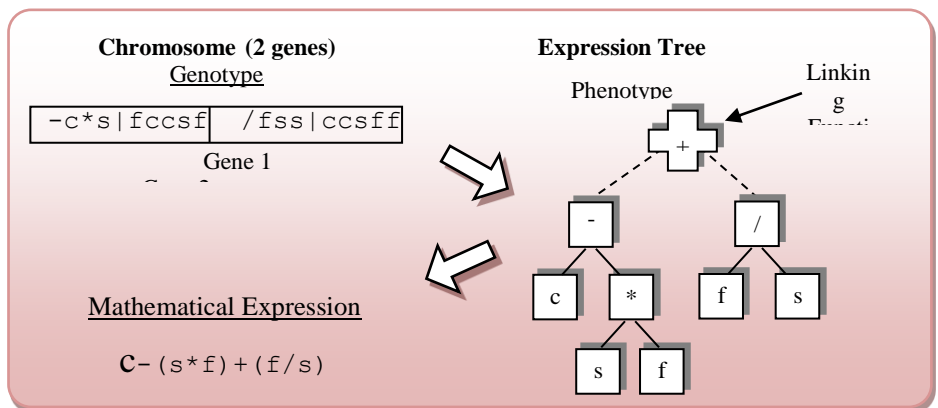

**Figure (3) Multi-gene Chromosomes with Linking Function**

It is reasonably familiar that GEP genes have non-coding regions downstream the termination point (Figure (4)). These regions are, in fact, the core of GEP and the process of evolution; they permit the modification of the genome by any genetic operator without restrictions, producing always syntactically correct programs [10]. This is the dominant difference between GEP and GP. When a mutation occurs in these regions, it is called neutral mutation, as it will have no effect on introducing variety. Also, these regions are ideal areas for chromosomes to split and crossover without interfering with the ORFs.

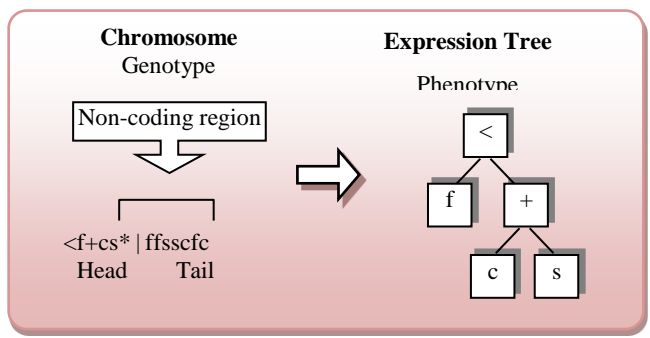

**Figure (4) The Chromosomal Non-coding Region in GEP**

Selected according to their fitness and the luck of the roulette wheel, chromosomes are subjected (by a certain probability) to change in their genetic values by one of the following GEP operators:
- Mutation.
- Transposition: IS (Insertion Sequence), RIS (Root Insertion Sequence) and GIS (Gene Insertion Sequence).
- Recombination: One-point recombination, Two-point recombination and Gene recombination.

## 6. The Proposed Approach:

In this work an attempt to solve the bin-packing problem is carried out with the use of GEP. As stated earlier, one-dimensional online problems are considered with the goal of building a system that can generate hyper-heuristics through evolution. These heuristics should be able to compete and improve their fitness to produce good solutions.

GEP is implemented with a function set including some or all of the following :{ + , - , * , / ,< } and a terminal set including :{ f: the sum of the pieces already in the bin, c: the bin capacity, and s: the size of the current piece}. Single-gene and multi-gene chromosomes are used each with varied gene length to show complexity of the resulting rules.

Figure (5) explains the strategy of filling the bins which is carried out in accordance to that used in Genetic Programming. This strategy is performed for each chromosome to test its efficiency and calculate its fitness value. After filling all the bins, the fitness function can be calculated and the evolutionary process is continued.

### 6.1. The Fitness Function:

Part of solving a problem efficiently is the design of fitness function. This largely influences the success of finding good and acceptable solutions to a problem. The objective must be correctly defined in order to make the system evolve in that direction.

If B is the number of bins used with (n) pieces each of size $S_k$ (for all k=0..n), pieces are to be packed into bins of capacity C, then the fitness measure is given by:

$$\text{Fitness} = B - \frac{\sum_{k=1}^{n} S_k}{C} \quad \text{………………………………………………………} (4)$$

The perfect result is gained when the fitness is zero, since the pieces are placed into the smallest possible number of bins. To distinguish illegal individuals that are far away from solving the problem, a high fitness value of 10000 is assigned to the function as compared to the range of fitness values that a legal solution can have [5].

### 6.2. Data sets:

The test data collection is available from Beasley's OR-Library [2], which is a collection of test data sets for a variety of OR problems. This test collection contains problems of two kinds that were generated and largely studied by Falkenauer [9]. For one-dimensional Bin-Packing Problems, there are 8 data files; their format is as follows:

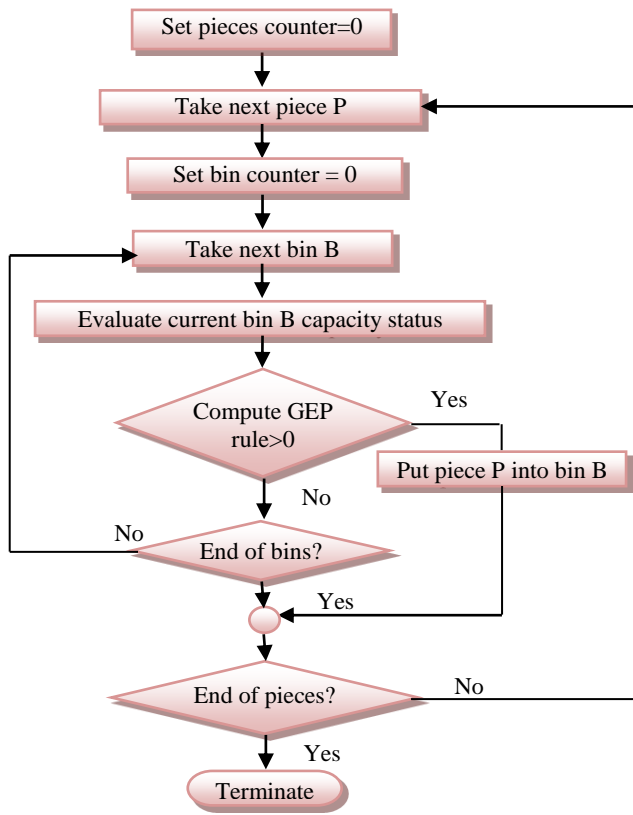

**Figure (5) Bin-Filling Strategy**

- No.P= Number of test problems
- For each test problem (P=1... No.P):
  ✓ Problem identifier,
  ✓ Bin capacity,
  ✓ Number of items (No.I),
  ✓ Number of bins in the current best known solution and
  ✓ For each item i (i=1... No.I): size of the item

In these files, two classes of bin-packing instances are found. In the first class, there are 4 files with problem identifiers beginning with ('u'). They consist of items of sizes uniformly distributed in (20,100) to be packed into bins of size 150. The second class, the rest of the files, also 4, problem identifiers begin with ('t'). These consist of 'triplets' of items from (25, 50) to be packed into bins of size 100.

This work is concerned with the first uniform class, the value for "Number of bins in the current best known solution" is the one found by Falkenauer [9] Except for problems u120_08, u120_19, u250_07, u250_12 and u250_13, this is also the smallest number of bins capable of accommodating all the items, so the value is the proven optimum [2].

## 7. Experimental Results

In order to evaluate the significance of this work, it is compared with that used by Burke, Hyde and Kendall [5] ; the same data set and similar parameter setting are employed to be used in the comparison. Four datasets were used from Falkenauer [9] as already stated; these four include varying number of pieces (from 120 to 1000 pieces) to be filled in the bins, which, in their turn, are fixed in their capacity to 150.

In addition, the impact or influence of functions is tested to specify the significance of each one and to choose the best to participate in the formation of rules.

Experiments carried out in this section are of twofold; first single-gene chromosomes are used with numerous gene lengths. Second, multi-gene chromosomes are investigated with different gene length and various gene numbers in the chromosome. The probability of success (Ps) notation is used to indicate success rates and all tests are evaluated by 100 identical run. The following subsections illustrate the various tests carried out in this work for both single and multi-gene chromosomes.

### 7.1. Using Single-gene Chromosomes

When faced with a problem to solve using heuristics, it is usually very important to know the components of the desired heuristic rule. In GP and GEP, the function and terminal sets given to the system fundamentally influence the success of that system. Sometimes, it is not clearly known in advance what the desired rule would look like, or even in some circumstances the output can be surprisingly different from the expected. That is why the components given to the system should be carefully examined and inspected. The following tests are carried out to find the right components used in forming the chromosomes. And since all three components of the terminal set have to be present in the rule to function correctly, only the function set should be examined.

This part of testing utilized single-gene chromosomes with gene length ranging from 5[h= 2 | t=3] to 15[h=7 | t=8]. The function set used contains functions that require

two arguments at most, so according to eq. (3) with (n=2) and the tail is computed as (head+1).

Table (1) shows the probabilities of genetic operators for the test carried out; the parameter settings used in this test are stated as follows:

- No of Generations= 50
- Population size= 150
- No. of Genes in each chromosome= 1
- Function Set : { + , - , *, / , <} (max_arg=2)
- Terminal Set:{ f: sum of the pieces already in the bin,
   c: bin capacity,
   s: size of the current piece.}

**Table (1) Probabilities of Genetic Operators for Tests**

| Mutation | Transposition | | | Recombination | | |
|---|---|---|---|---|---|---|
| | *IS* | *RIS* | *GIS* | *One* | *Two* | *Gene* |
| 0.05 | 0.3 | 0.2 | - | 0.4 | 0.3 | - |

To show the impact of each function in the set, each one is separately given twice the chance of the others to be selected to contribute in the gene of initial populations and also twice the chance to be chosen for mutation or crossover operators throughout the evolution.

The test is done by using the function set of {+,-,*,/,<} , the results are shown in Table (2), where each column is a set of runs with its label function being doubled to increase it chance in being picked up. The outcome shows how *subtraction* (-) surpasses other functions in its effect; this is due to the nature of the problem where subtraction has a big role in fitting pieces in bins. In the second place comes *addition* (+), then the *less than* (<) function, after that comes *division* (/), and last come *multiplication* (*) with the least effect as the Ps apparently dropped down.

**Table (2) Ps for Function Impacts**

| Length {h|t} | Ps with function set {+,-,*,/,<} | | | | |
|---|---|---|---|---|---|
| | - | + | < | / | * |
| 5  {2|3} | **80** | 60 | 65 | 65 | 48 |
| 7  {3|4} | **60** | 54 | 51 | 51 | 41 |
| 9  {4|5} | **53** | 68 | 52 | 54 | 40 |
| 11{5|6} | **65** | 53 | 43 | 42 | 39 |
| 13 {6|7} | **75** | 63 | 63 | 60 | 43 |
| 15 {7|8} | **71** | 57 | 62 | 49 | 48 |
| **Average Ps** | **67.33** | 59.17 | 56 | 53.5 | 43.17 |

*Applying Gene Expression Programming for …*

Another test is conducted this time with an attempt to eliminate (*). Using three different function sets, the results are stated in Table (3). The first and second columns compare between success rates using (*) and (/) along with other functions and the comparison ends up in favor of (/). The third separated column shows success rates using both (*) and (/) with other functions, and although the results are a little better than that in the second column but it is still not better than that of the first.

Studying the nature of the problem and desired rules in light of these results leads to the conclusion that (*) does not have a useful role in solving the problem. When looking at the obtained rules in runs of column 2 of Table (3), multiplication either appear in the non-coding region or have a pour effect like multiplying by -1 as shown in figure (6).

**Table (3) Ps using Single-Gene Chromosomes**

| Length {h|t} | Ps with 4 functions | | Ps with 5 functions |
|---|---|---|---|
| | +,-,/,< | +,-,*,< | +,-,*,/,< |
| 5  {2|3} | **68** | 61 | 63 |
| 7  {3|4} | **71** | 51 | 58 |
| 9  {4|5} | **57** | 48 | 64 |
| 11{5|6} | **67** | 45 | 54 |
| 13 {6|7} | **74** | 57 | 63 |
| 15 {7|8} | **64** | 51 | 61 |
| **Average Ps** | **66.83** | 52.17 | 60.5 |

So, according to this test, multiplication was moved out of the competition leaving other functions to fight for survival. Function impact Test was performed again this time without (*). Table (4) shows how the ranking of functions gained in Table (2) was confirmed, the average Ps improved from 67.33 to 80.5 for subtraction. Other functions' success rates also improved indicating a better overall performance for the entire system.

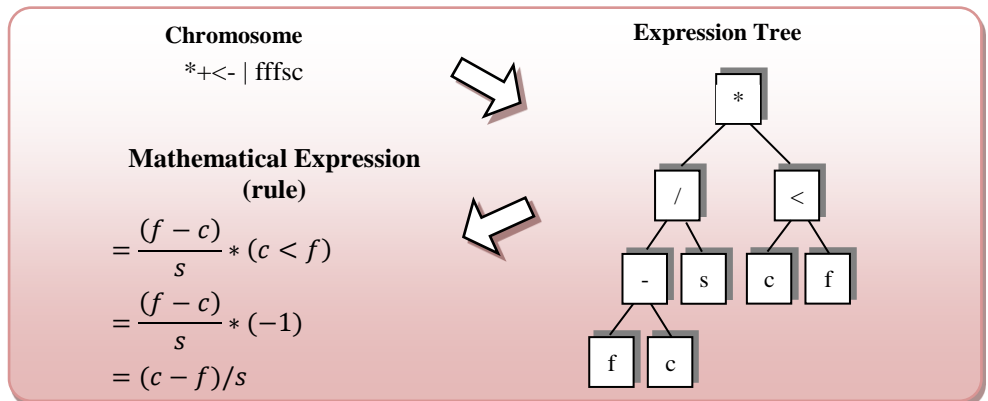

**Figure (6) An Example from Test Results**

**Table (4) Ps for Single-Gene Chromosome using 5 Varied Functions**

| Length {h|t} | Ps with function set {+,-,/,<} | | | |
|---|---|---|---|---|
| | - | + | < | / |
| 5  {2|3} | **87** | 71 | 70 | 72 |
| 7  {3|4} | **80** | 77 | 68 | 66 |
| 9  {4|5} | **72** | 53 | 56 | 51 |
| 11{5|6} | **78** | 54 | 57 | 55 |
| 13 {6|7} | **79** | 77 | 67 | 69 |
| 15 {7|8} | **87** | 61 | 69 | 60 |
| **Average Ps** | **80.5** | 65.5 | 64.5 | 62.17 |

### 7.2. Using Multi-Gene Chromosome

In this subsection, multi-gene chromosomes are investigated taking into account both the number of genes in each chromosome and the length of each gene. To show the influence of gene numbers in the chromosomes on success rates, two, three, and four genes are used in the chromosome; this is done in each test along with various gene lengths. Table (5) shows the probabilities of genetic operators used in all the tests in this subsection.

The functions set of { + , - , / , < } is used in this test as a consequence of the results obtained previously from tests carried out in subsection (7-1). Tests are carried out using all functions in the function set as a linking function. Table (6) shows the success rates by using (-) for linking genes in the chromosomes, and apparently gave the best results.

After that (<) is used to link genes, results in Table (7) are comparatively good. Next (/) is employed in Table (8) with less success rates, and at the end (+) in Table (9) gave the poorest success rate among all functions. Logically (*) was not used in this test also due to the fact that it has a bad impact on success rates.

**Table (5) Probabilities of genetic operators for tests in this subsection**

| Mutation | Transposition | | | Recombination | | |
|---|---|---|---|---|---|---|
| | *IS* | *RIS* | *GIS* | *One* | *Two* | *Gene* |
| 0.05 | *0.3* | *0.3* | *0.2* | *0.2* | *0.2* | *0.1* |

**Table (6) Ps using *subtraction* in linking the genes in the chromosomes**

| Length {h\|t} | Two Genes | Three Genes | Four Genes |
|---|---|---|---|
| 5 {2\|3} | 87 | ***98*** | 98 |
| 7 {3\|4} | 80 | ***100*** | 99 |
| 9 {4\|5} | 92 | ***94*** | 95 |
| 11{5\|6} | 89 | ***100*** | 98 |
| 13 {6\|7} | 84 | ***98*** | 94 |
| 15 {7\|8} | 93 | ***97*** | 96 |
| Average Ps | 87.5 | ***97.83*** | 96.67 |

**Table (7) Ps using *less than* in linking the genes in the chromosomes**

| Length {h\|t} | Two Genes | Three Genes | Four Genes |
|---|---|---|---|
| 5 {2\|3} | 86 | 64 | 49 |
| 7 {3\|4} | 87 | 58 | 43 |
| 9 {4\|5} | 87 | 56 | 39 |
| 11{5\|6} | 83 | 56 | 49 |
| 13 {6\|7} | 82 | 65 | 54 |
| 15 {7\|8} | 87 | 60 | 53 |
| Average Ps | **85.33** | 59.83 | 47.83 |

**Table (8) Ps using *division* in linking the genes in the chromosomes**

| Length {h\|t} | Two Genes | Three Genes | Four Genes |
|---|---|---|---|
| 5 {2\|3} | 62 | 58 | 33 |
| 7 {3\|4} | 60 | 44 | 42 |
| 9 {4\|5} | 61 | 51 | 35 |
| 11{5\|6} | 56 | 49 | 42 |
| 13 {6\|7} | 68 | 55 | 43 |
| 15 {7\|8} | 61 | 53 | 45 |
| Average Ps | **61.33** | 51.67 | 40 |

**Table (9) Ps using *addition* in linking the genes in the chromosomes**

| Length {h\|t} | Two Genes | Three Genes | Four Genes |
|---|---|---|---|
| 5 {2\|3} | 51 | 23 | 11 |
| 7 {3\|4} | 38 | 46 | 25 |
| 9 {4\|5} | 49 | 32 | 31 |
| 11 {5\|6} | 53 | 55 | 22 |
| 13 {6\|7} | 55 | 47 | 34 |
| 15 {7\|8} | 56 | 45 | 33 |
| **Average Ps** | **50.33** | 41.33 | 26 |

Figure (7) gives some examples from test runs for each linking function, also shown in the figure how each chromosome is encoded and then converted to a heuristic rule at the implementation of the fitness function.

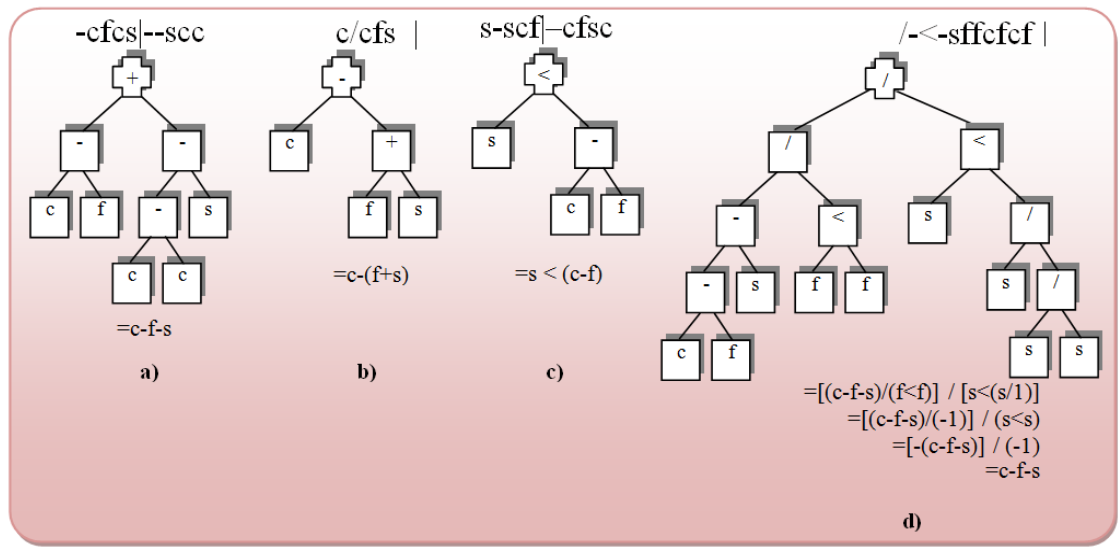

**Figure (7) Solution Examples using different linking functions**

**a) Using addition . b) Using subtraction. c) Using less than. d) Using division.**

### 7.3. Number of Generations:

In all the previous tests, the number of generations was fixed to 50 generations. This was done due to the comparisons made with Burke in [5]. In this third test, the number of generations is inspected. Using single-gene chromosomes with function set of {+,-, /,<}, head of genes spans from 2 to 7. The number of generations is varied from 30 to 50, increasing five generations for each try. Figure (8) shows the average of Probability of success for each number of generations used. As seen from the results in the above figure, 35 generations were able to give the best Ps among all others.

When using 40 generations a special case arises, as Ps evidently drop down because of the fact that 35 generations are the required number to evolve appropriately; once extra generations are added, the system crashes and cannot adapted to that additional evolutionary space. Adding more generations gave the evolved system sufficient time to improve again.

### 7.4. Comparison with Previous Work:

When comparing this work with that of Burke [5]; many issues can be concluded as stated in Table (10). The accuracy of solutions is not the issue of comparison; the optimal solution is known and given in [9], the evolved rules gave the same results, but the main issue of concern is in terms of the genetic parameters, computational effort, and memory storage.

The required number of generations is less, the size of the population is much better; a population of 1000 individuals is very huge, while 150 is neither too small nor too large and requires less memory storage.

The size of the population is very critical to the process of evolution and although it is commonly thought that large populations evolve more rapidly than smaller ones, because of their increased rate of mutations, large populations evolve mostly deterministically and often become trapped on local fitness peaks, smaller populations can follow more stochastic evolutionary paths and thus locate higher fitness peaks [17].

In practical applications, large populations require more computational effort for moving to the next generation. This has the effect on the model when dynamic optimization problems are studied. A population of size *n* requires $O(n)$ computations to advance one generation [24].

**Table (10) Difference between Current and Previous Work**

|  | **Current work** | **Previous work [5]** |
|---|---|---|
| **Population size** | 150 | 1000 |
| **Generations** | 35 | 50 |
| **Maximum depth of initial trees** | No need for limits (max 3) * | Needs a Limited (max set to 4) |

 * (when max-arg=2 and max gene length=15 the tree cannot be more than 3 levels deep in single-gene Chromosomes)

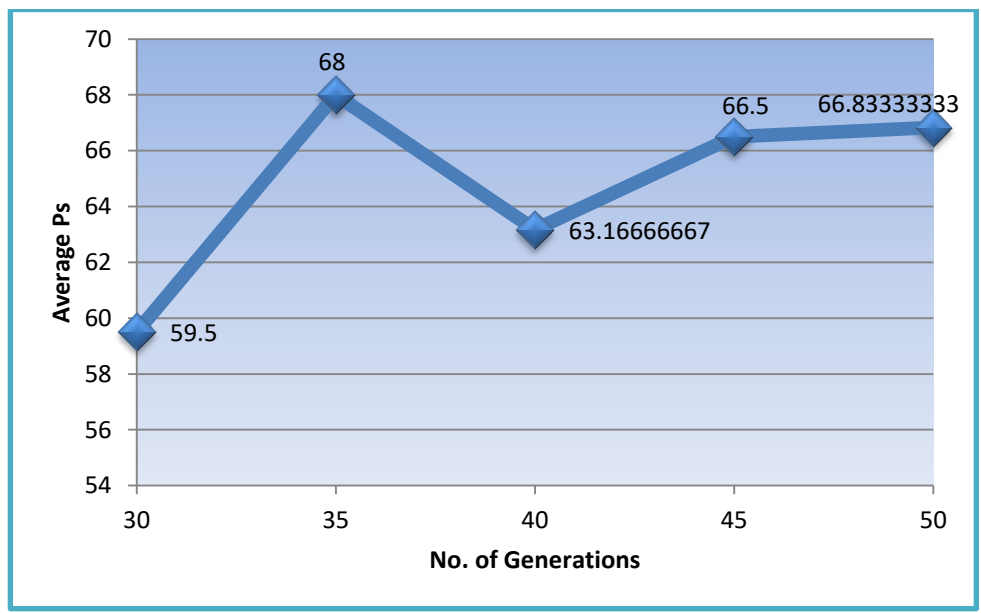

**Figure (8) Single-Gene Chromosomes with Various Generations**

In general, the use of a genotype/phenotype system such as GEP instead of GP has the following advantages:

1- GEP is easier to program than GP; genetic operations are done on linear chromosomes instead of non-linear tree structures.
2- In GP the length of the chromosome is not predictable and can easily cause system overflow problems, especially in initiating the first population or after applying the crossover operator.
3- GEP utilizes chromosomes with multiple genes in comparison to the limited single-gene chromosomes of GP.
4- Bloating [26] (the uncontrolled and unbounded growth of individuals in the population, i.e., programs become too large) is one of the fundamental problems of GP.
5- Crossover and mutation problems, as they can result in erroneous genes when using GP.
6- Very good heuristics can be discovered by using smaller size populations and fewer generations than that used in GP.

It is clear to state that the application of GEP to Bin-Packing has the following advantages:
1- Interesting heuristics can be found to solve the problem.
2- The use of the learning idea imbedded in the nature of GEP.
3- Parallelism is achieved when evolving many rules at the same time in the population.
4- Crossover operations can sometimes work as an exchange procedure of useful rule components between the different fit heuristics to increase their efficiency.
5- The complexity of the rules is controlled through the length of the chromosome and number of gene in the chromosome.
6- As a further advantage of using this approach is that any rule can be expressed as a GEP chromosome and injected into the population for testing, as a matter of fact, a whole population can be initiated with predefined heuristic rules for testing.

## 8. Conclusions and Further Work

The utilization of GEP to Bin-Packing problems has many advantages, mainly finding good heuristics to solve the problem, and controlling the complexity of these heuristics by varying the chromosome's length or by changing the number of gene in each chromosome. Also, the use of learning imbedded in the nature of GEP is very useful. Another benefit is gained through performing a parallelism when evolving many rules at the same time in the population. The crossover operator work sometimes as an exchange procedure of useful rule components between the different fit heuristics to increase their efficiency. .An additional advantage is testing; as different rules can be encoded in chromosomes and inserted into populations for testing, or even an entire population can be filled with certain rules to undergo evolution and verify their fitness.

While using GEP instead of GP solves many issues such as the difficulty of programming on tree structures. Length of genes and chromosomes is no longer an issue neither in first populations nor after applying crossover operators (and no Bloating can occur). Another issue is multiple genes and their role in producing more complex rules. The discovery of very good heuristics rules using smaller size populations and fewer generations. Crossover and mutation problems are eliminated and cannot result in erroneous genes

As for a further recommendation, GEP algorithm can be studied and utilized for the application of two- , three- and multi- dimensional problems. It can also be easily modified to include other constraints. In addition, bin-packing variations can be considered, such as sequential bin-packing problems or on-line bin-packing problems with rejection.

A potential area for future work can target the range of functions and terminals given to the process of evolution, the introduction of any new useful component could have an immense impact on increasing the complexity of rules to be evolved.

GEP can be further tested over more bin-packing benchmark datasets to investigate the efficiency of the procedure under different circumstances. The second class of Falkenauer datasets can be put to the test, that consist of 'triplets' of items from (25, 50) to be packed into bins of size 100.

*Applying Gene Expression Programming for ...*